\documentclass[runningheads]{llncs}

\input{psfig.sty}
\usepackage{graphicx}
\usepackage{caption}
\usepackage{subcaption}
\usepackage{subfig}
\usepackage{amsmath}
\usepackage{algorithm}
\begin{document}

\pagestyle{headings}

\mainmatter

\title{Fast Mesh-Based Medical Image Registration}

\author{Ahmadreza Baghaie\inst{1} \and Zeyun Yu\inst{2}\thanks{corresponding author: yuz@uwm.edu} \and Roshan M. D'souza\inst{3} }
\authorrunning{A. Baghaie et al.}

\institute{Department of Electrical Engineering,University of Wisconsin-Milwaukee,WI 53211 \and Department of Computer Science,University of Wisconsin-Milwaukee,WI 53211 \and Department of Mechanical Engineering,University of Wisconsin-Milwaukee,WI 53211}

\titlerunning{Lecture Notes in Computer Science}

\maketitle

\begin{abstract}
In this paper a fast triangular mesh based registration method is proposed. Having \textit{Template} and \textit{Reference} images as inputs, the template image is triangulated using a content adaptive mesh generation algorithm. Considering the pixel values at mesh nodes, interpolated using spline interpolation method for both of the images, the energy functional needed for image registration is minimized. The minimization process was achieved using a mesh based discretization of the distance measure and regularization term which resulted in a sparse system of linear equations, which due to the smaller size in comparison to the pixel-wise registration method, can be solved directly. Mean Squared Difference (MSD) is used as a metric for evaluating the results. Using the mesh based technique, higher speed was achieved compared to pixel-based curvature registration technique with fast DCT solver. The implementation was done in MATLAB without any specific optimization. Higher speeds can be achieved using C/C++ implementations.

\end{abstract}

\section{Introduction}
\textit{Medical Image Registration} is an active area in the field of image processing with applications ranging from image mosaicing in retinal images \cite{1} to slice interpolation \cite{2} etc.  Image registration problems can be categorized into four major categories: \textit{multi-view analysis, multi-temporal analysis, multi-modal analysis} and \textit{scene to model registration} \cite{3}. Generally speaking, having two images, \textit{Reference} (Re) and \textit{Template} (Te), the image registration problem is to find a \textit{valid} and \textit{optimal} spatial or geometrical transformation between the two input images. In the process, the pixel values of template image won't change and only the locations will be altered. \
Being more focused on the implementation aspects of image registration, two major classes of methods can be considered: \textit{Parametric} and \textit{Non-parametric} image registration methods \cite{4}. Parametric registration methods are based on a finite set of parameters or image features. These methods include rigid registration, affine registration, land-mark based registration, principal axes-based registration, FFT-based registration, optimal linear registration and spline registration. Unlike parametric registration methods,  the non-parametric methods are not limited to a finite set of parameters. Diffusion registration, fluid registration, curvature registration and elastic registration \cite{5} are a few examples of this class of methods.

In general, image registration is considered as an \textit{ill-posed inverse problem}. Therefore the process of solving this problem consists of three component \cite{6}: 1) a deformation model, 2) an objective function to be optimized and 3) an optimization method. A general objective function can be defined as:
\begin{equation}
 E[u]=D[Re,Te \circ u]+\alpha S[u] 
\end{equation}
The left hand side of the equation is the energy or objective function which needs to be optimized; $ u $ is the displacement field. On the right, we have two terms: $ D $ and $ S $. The first term is called \textit{(dis)similarity measure} or \textit{distance measure} which acts as the matching criterion between the reference image and the deformed template image, $ u $ being the displacement field. Depending on the choice of this term, the objective function needs to be either minimized or maximized. Some common examples of the distance or similarity measure are Mutual Information (MI), Normalized MI (NMI), Cross-Correlation (CC), Normalized CC, Sum of Squared Differences (SSD) etc. which are widely used in the literature in this area \cite{7}. The second term, $ S $ is the \textit{regularization term} which imposes additional constraints on the deformation. Due to ill-posedness of image registration, this term is needed to be able to have reasonable transformations. This term dictates the validness of transformation. The parameter $ \alpha $ is a weight which indicates the amount of regularization. 

Other than the obvious way of assessing an image registration method which is the similarity of the final deformed template image to the reference image, the computational complexity of the method is also a major factor. This becomes more important in case of non-rigid image deformation methods in which the deformation is local rather than global. The main reason for this is due to the size of the images, which results in a very big number of degrees of freedom in the optimization process. This is more obvious in case of non-parametric image registration techniques. One obvious solution to remedy this problem is using computers with higher computational power. Using parallel processors such as Graphics Processing Units (GPUs) is an example which has been widely considered in the past few years \cite{8}. 

From an algorithmic point of view, using  multi-resolution techniques can also be considered \cite{9} which start from a very coarse grid to capture the larger deformations in the images to be registered and then move to much finer grids to capture the smaller deformations. Usually, this is done on a uniform square grid which means that the sampling will take place uniformly. In other words, even though the objects and image features are scattered randomly in the images to be registered, multi-resolution based methods do not take this fact into consideration. Adaptive griding using Octrees has been used to address this issue \cite{10}. However, square grids cannot accurately match feature boundaries that are typically curvilinear. Using triangular meshes enables better representation of curvilinear image feature boundaries. This representation also allows for easy reconstruction of final results as well as the computation of the final optimized displacement. Furthermore, content-based adaptive meshing greatly reduces the computational complexity of the registration process. Of course, this will have its own implications for optimizing the energy functional of the method which will be discussed more in later sections. 

There have been a few previous works in this area, usually considering the problem of image registration as a \textit{Finite Element Method} (FEM) problem \cite{11}. But they usually need additional information about the physical properties of the underlying structure. Also the process is not completely mesh based. In \cite {11} each update of the displacement field in each iteration is followed by a re-sampling to the regular image grid, computational procedures to find the update terms for the next update and then re-sampling again to the mesh triangular grid. 

In this paper, a new mesh based image registration technique will be introduced which takes advantage of mesh-based operators without the need to switch between regular image grid and triangular mesh grid. Taking the reference and template images as the inputs and some needed parameters, the proposed method can achieve excellent accuracy and higher speed compared to regular pixel-based registration methods. Even though a uniform initial mesh can be considered, an initial mesh is created for the template image using a content adaptive mesh generation method which perfectly matches the image's edges and features. The energy functional is minimized next and the final displacement field is reconstructed. Finally, MSD will be used in order to assess the performance of our registration technique.

 The paper is organized as follows: Section 2 contains comprehensive details about the proposed method. Results and discussions are provided in Section 3. Section 4 concludes the paper.

\section{Proposed Method}
Taking the general notation as most of the papers in this field, assuming \textit{Template} (\textit{Te}) and \textit{Reference} (\textit{Re}) images as inputs, the goal of image registration is to find a valid and optimal geometrical transformation to be applied on \textit{Te} to become more similar to \textit{Re} according to some similarity measure. Therefore, the process can be formulated as an optimization problem which tries to optimize some energy functional that can be defined as in (1). Due to the random distribution of image features, we propose to use a sparser representation of both input images using a content adaptive mesh generator such as the one described in \cite{12}. 

\subsection{Formulation Of The Mesh-Based Image Registration Method}
Assume $ Te $ and $ Re $ as input images with the same size, and a set of triangles defined on the template image represented by $ (V, T) $, where $ V $ is a $ n_V\times2 $ matrix containing the coordinates of $ n_V $ mesh nodes or vertices and $ T $ is a $ n_T\times3 $ matrix, each line containing the indexes of nodes creating each one of the $ n_T $ triangles. Note that $ Te(V) $ represents a vector with constant values which only the locations of its values change in the process of optimization. $ Re $ represents a continuous domain of $ X\in\Omega $, hence $ Re(V)=Re(X)|_{X=V} $. Also, it should be mentioned that the set of triangles covers the template image domain $ \Omega $. Here a slightly different approach will be considered in which instead of applying the smoothing at the same time as update of the displacement filed, this will be done after each iteration using a diffusion process which will be described later in more details. The energy functional is therefore define as follows: 
\begin{equation}
E[u(V)]=D[Re(X)|_{X=V+u}, Te(V)\circ u(V)]
\end{equation}
where $ E $ and $ D $ represent the energy functional and the distance measure respectively. Also, the $ \circ $ operator is defined as:
\begin{equation}
Te(V)\circ u(V)=Te(V+u(V))
\end{equation}

For simplicity of representation, and since it is obvious that the method is applied to mesh nodes, from now on, the notation of a function $ f $ of variable $ V $ which has a general form of $ f(V) $ will be reduced to just $ f $. As mentioned before, several distance or similarity measures can be found in the literature, each having its pros and cons and being suitable for different problems encountered in the image registration. Here, the Sum of Squared Differences (SSD) is used which can be defined as follows:
\begin{equation}
\begin{split}
D(Re(X)|_{X=V+u}, Te \circ u)= \frac{1}{2}||Re(X)|_{X=V+u}-Te \circ u||^2\\
=\frac{1}{2}\sum _{i=1:n_V} (Te(V_i) \circ u(V_i) - Re(X_i)|_{X_i=V_i+u_i})^2
\end{split}
\end{equation}
where the last summation is computed over all of the mesh nodes. Minimizing the energy functional and updation of the displacement field can be done considering a gradient descent approach:
\begin{equation}
u_0 ^{k+1}=u_1 ^k - \tau \bigtriangledown _{u_1^k} E[u_1^k]
\end{equation}
where $ \tau $ is the step size (here 0.005) and $ \bigtriangledown _{u_1^k} $ is the gradient operator with respect to variable $ u_1^k $. The gradient of the energy functional is computed by taking the Gateaux derivative of the distance measure which results in:
\begin{equation}
\begin{split}
\bigtriangledown _{u_1^k} E[u_1^k]=\bigtriangledown _{u_1^k} D \hspace{50 mm} \\=(Te(V+u_1^k)-Re(X)|_{X=V+u}).\bigtriangledown _{u_1^k} Te(V+u_1^k)
\end{split}
\end{equation}
where $ \bigtriangledown _{u_1^k} Te(V+u_1^k) $ needs to be computed on mesh nodes. Further explanation will be given in Section 2.1. 
The reason behind using two different subscripts (0 an 1) in (5) is because of the fact that this displacement function needs to be smoothed to ensure regularized displacements in the image domain. For smoothing the displacements on the mesh, a diffusion process needs to be solved on the mesh nodes. This diffusion process can be modeled as follows:
\begin{equation}
\frac{\partial u_0^{k+1}}{\partial t}=\lambda \bigtriangleup u_0^{k+1}
\end{equation}
where $ \bigtriangleup  $ represents the Laplacian operator on mesh nodes. This diffusion process is solved using a forward difference time-stepping approach. Without loss of generality and to reduce the confusion with the gradient descent method's step size, here the time step will be considered as 1. Hence from (7):
\begin{equation}
u_1^{k+1}=u_0^{k+1}+\lambda \bigtriangleup u_0^{k+1}
\end{equation}
where $ 0< \lambda <1 $ is the smoothing parameter defined by the user (here 0.8). 
Further simplification will be done in Section 2.3.

\subsection{Discretization of Gradient On A Triangular Mesh}
Consider node $ V_i $ and its 1-ring ($ N_1 $) neighbor nodes. Approximation of the gradient of a function $ f $ on the location of node  $ V_i $ can be achieved using linear interpolation of the function $ f $ over the surface created by this region. Assuming triangle $ T_j $ created by nodes $ [V_i V_j V_k] $ as one of the triangles surrounding $ V_i $, the approximation of the gradient on $ T_j $ will be:
\begin{equation}
\begin{split}
\bigtriangledown f_{T_j} = \frac{1}{4 A_j^2} \bigg( f_i [(\overrightarrow{V_{ij}}, \overrightarrow{V_{jk}})(V_k - V_i) + (\overrightarrow{V_{ik}}, \overrightarrow{V_{kj}})(V_j - V_i)]\\
+ f_j [(\overrightarrow{V_{ji}}, \overrightarrow{V_{ik}})(V_k - V_j) + (\overrightarrow{V_{jk}}, \overrightarrow{V_{ki}})(V_i - V_j)]\\
+ f_k [(\overrightarrow{V_{kj}}, \overrightarrow{V_{ji}})(V_i - V_k) + (\overrightarrow{V_{ki}}, \overrightarrow{V_{ij}})(V_j - V_k)] \bigg)
\end{split}
\end{equation}
where $ f_i $ is the function value on node $ V_i $, $ A_j $ is the area of the triangle $ T_j $, $ \overrightarrow{V_{ij}} $ is the vector connecting nodes $ i $ and $ j $ and $ (\overrightarrow{a}, \overrightarrow{b}) $ gives the \textit{dot} product of vectors $ \overrightarrow{a} $ and $ \overrightarrow{b} $. 
Having the approximation of the gradient on surrounding triangles, the approximate gradient for node $ V_i $ can be computed as follows:
\begin{equation}
\bigtriangledown f(V_i) = \frac{1}{A(V_i)} \sum _{j \in N_1 (i)} A_j \bigtriangledown f_{T_j}
\end{equation}
where $ A(V_i)= \sum _{j \in N_1 (i)} A_j $. For a complete analysis on the approximation error the reader is referred to \cite{13}. The areas of triangles should be computed at the beginning of each iteration. 

\subsection{Diffusion-Based Smoothing Of Displacement}
Taking the same approach as \cite{14}, the Laplacian operator on a mesh can be approximated by the so-called \textit{umbrella} operator on each node as follows:
\begin{equation}
\bigtriangleup u(V_i) =\frac{1}{m_i} \sum _{j \in N_1 (i)} u(V_j)-u(V_i)
\end{equation}
where $ m_i $ is the valence (number of 1-ring neighbors) of node $ V_i $. This operator can be defined in a matrix form as follows:
\begin{equation}
\bigtriangleup u=(A^{Lap}-I)u
\end{equation}
where $ I $ is the identity matrix and $ A^{Lap} $ is a sparse $ n_V \times n_V $ matrix which its non-zero elements are defined as follows:
\begin{equation}
A_ {ij} ^{Lap} = \frac{1}{m_i}, \ for \ all \  j \in N_1(Vi)
\end{equation}

Considering (8) and (13) together with a few manipulations, the diffusion process can be simplified as a weighted average of the displacements of the 1-ring neighborhood of each node:
\begin{equation}
u_1^{k+1}=\big((1-\lambda)I+\lambda A^{Lap}\big)u_0^{k+1}
\end{equation}

The above equation can be applied iteratively for further smoothness of the displacement field on the mesh nodes. Here, only one iteration of smoothing is applied. The overall algorithm for mesh-based registration is illustrated in Algorithm 1.

\begin{algorithm}
\caption{Fast Mesh-Based Image Registration}
\textbf{Inputs}: $ Re $, $ Te $, $ (V, T) $ defined on the template image, $\lambda$, $\tau$ ;\\
\textbf{Pre-Computation}: $ N_1 $ and neighbor triangles for each mesh node, $ A^{Lap} $;\\
\\
For $ k=1\rightarrow convergence $\\
{\{
\begin{itemize}
\item \textit{Update}:
\begin{itemize}
\item {$ E[u]=D(Re(X)|_{X=V+u}, Te\circ u) $}
\item {$ \bigtriangledown _{u_1^k} E[u_1^k]$}
\item {$ u_0 ^{k+1}=u_1 ^k - \tau \bigtriangledown _{u_1^k} E[u_1^k] $}
\end{itemize}
\item \textit{Smoothing}:
\begin{itemize}
\item {$ u_1^{k+1}=\big((1-\lambda)I+\lambda A^{Lap}\big)u_0^{k+1} $}
\end{itemize}
\end{itemize}
}\}
\end{algorithm}

\section{Results And Discussion}
\subsection{Content Adaptive Mesh Generation}
For generating the content adaptive mesh needed for our algorithm, the method proposed by Ming et al. \cite{12} is used. Based on the discussion given in this paper, the main difference between various mesh generating methods rises from different methods used for node placement. In some, the nodes are placed based on feature points in images, while in others, this process is done iteratively,  either starting from a coarse mesh and adding new nodes or starting from a dense mesh and removing redundant nodes. The method consists of several steps as follows:
\begin{enumerate}
\item Node generation:
	\begin{itemize}
	\item Canny sample points;
	\item Halftoning sample points;
	\item Uniform sample points.
	\end{itemize}
\item Mesh generation via Delaunay triangulation;
\item Image-based mesh smoothing:
	\begin{itemize}
	\item Image-based Centroid Voronoi Tessellations (CVT) mesh smoothing;
	\item Image-based Optimal Delaunay Triangulations (ODT) mesh smoothing;
	\item Edge flipping.
\end{itemize}	 
\end{enumerate} 

The result of this method is a high quality content adapted triangular mesh which is matched accurately with image features and edges. Fig. 1 displays an example of content adapted mesh generated for a brain cross-section. Image registration results using the proposed method is given in the following sections.

\begin{figure}
	\centering
	\begin{subfigure} [b]{0.35\textwidth}
	\includegraphics[scale=.25]{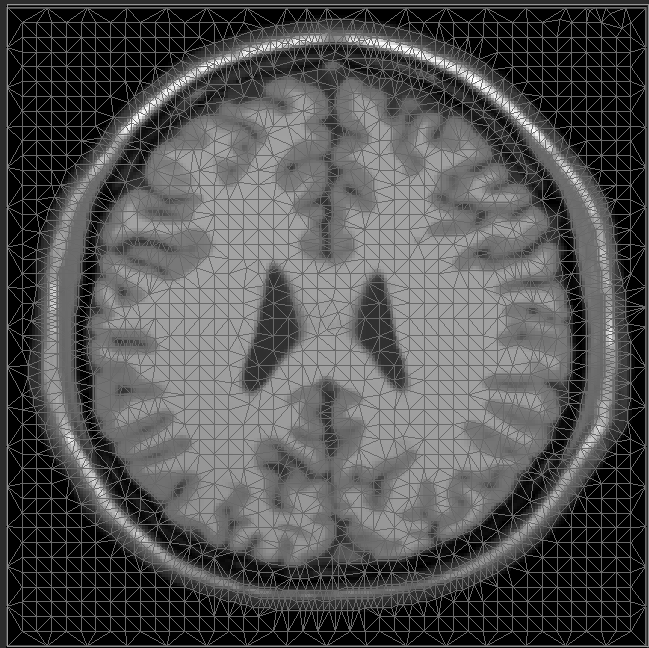}
	\caption{}
	\end{subfigure}	
	\begin{subfigure} [b]{0.35\textwidth}
	\includegraphics[scale=.32]{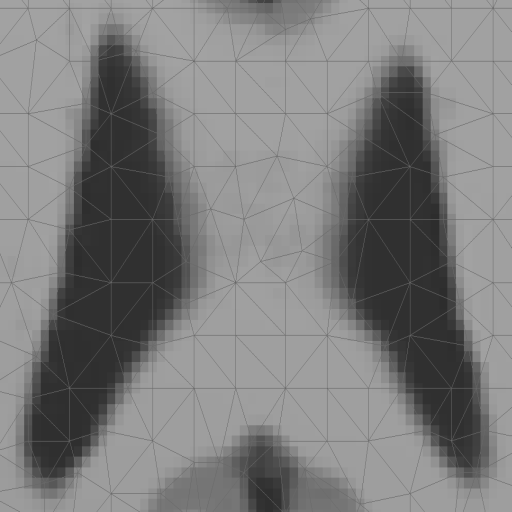}
	\caption{}
	\end{subfigure}
\caption {Example of content adaptive mesh generation}
\end{figure}

\subsection{Example 1- Brain CT Images}
For the first example, a pair of brain images are considered which are displayed in Fig. 2. Comparing the reference and template images reveals a rigid transformation as well as a non-rigid transformation in the center between the two. Using a content adapted mesh with 5406 nodes and 10744 triangles, the registration is done. The average time for each iteration is about 156 ms for these images. The computed displacement fields as well as the registered image and the difference image after registration can be seen in Fig. 3. The MSDs before and after registration are 271.8 and 77.3 respectively.

\begin{figure}
	\centering
	\begin{subfigure} [b]{0.3\textwidth}
	\includegraphics[scale=.45]{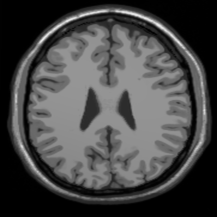}
	\caption{}
	\end{subfigure}	
	\begin{subfigure} [b]{0.3\textwidth}
	\includegraphics[scale=.45]{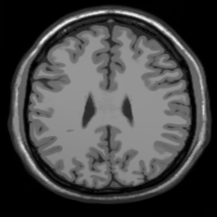}
	\caption{}
	\end{subfigure}
	\begin{subfigure} [b]{0.3\textwidth}
	\includegraphics[scale=.45]{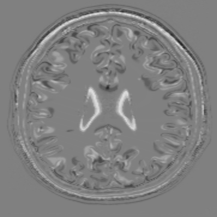}
	\caption{}
	\end{subfigure}	
\caption {(a) Template image, (b) Reference image, (c) Difference image}
\end{figure}

\subsection{Example 2- Brain CT Database}
In this section, a complete database of brain CT images are considered. The database contains 80 images,  each of the size of $ 512\times 512 $ pixels. Using the content adaptive mesh generation method, a mesh generated for each image in the database and then used for registration of consecutive slices in the database. Each mesh contains approximately 3300 nodes and 6700 triangles. For better comparison of the speed of the proposed method with pixel-based registration, an implementation of the curvature-based registration method \cite{5} has been used. This implementation takes advantage of a fast Discrete Cosine Transform (DCT) solver. Both of the methods are implemented and tested on MATLAB without any specific optimization and the process of optimization is terminated after 100 iterations. For the pixel-based curvature registration method, the DCT solver is implemented using the embedded DCT function in MATLAB which uses a C implementation, therefore is very fast and optimized while in the implementation of our method, the solver is implemented using in MATLAB scripts by the authors. However, the proposed method performs faster. Using a MEX or C implementation of the proposed method, higher speeds can be achieved. Table 1 summarizes the computational time of these two methods, implemented on a desktop computer with an Intel Core i7 3.5 GHz CPU and 6 GB of RAM, as well as the mean MSD error of the methds.

\begin{table}[!t]
\renewcommand{\arraystretch}{1.3}
\caption{Computational time and mean MSD error for pixel-based and mesh-based registration methods}
\centering
\begin{tabular}{c c c}
\hline
\bfseries   &   \bfseries Pixel-based Method  & \bfseries Mesh-based Method\\
\hline\hline\
\bfseries  Mean MSD &    116.66  &  108.91\\
\hline
\bfseries  CPU Time &    1534 sec  &  1320 sec\\
\hline\hline\
\end{tabular}
\end{table}

\begin{figure}
	\centering
	\begin{subfigure} [b]{.69\textwidth}
	\includegraphics[scale=.3]{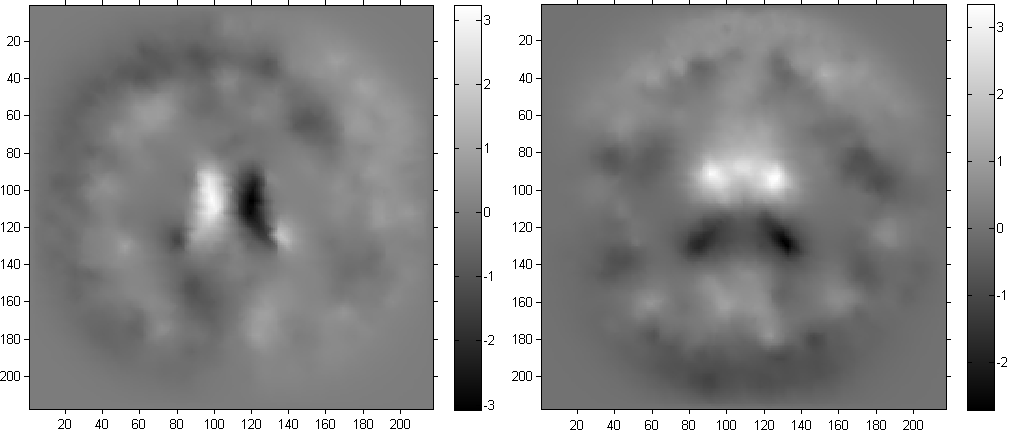}
	\caption{}
	\end{subfigure}	\\
	\begin{subfigure} [b]{0.3\textwidth}
	\includegraphics[scale=.4]{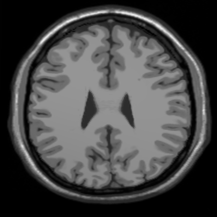}
	\caption{}
	\end{subfigure}	
	\begin{subfigure} [b]{0.3\textwidth}
	\includegraphics[scale=.4]{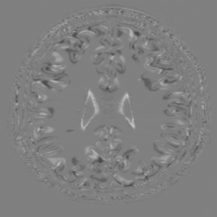}
	\caption{}
	\end{subfigure}	
\caption {(a) Displacement fields in horizontal and vertical directions, (b) Registered image, (c) Difference image after registration}
\end{figure}

\section{Conclusion}
In this paper, a new efficient triangular mesh-based image registration technique is introduced. Table 1 concludes the results of comparison between pixel-based curvature registration method with DCT solver \cite{5} with the proposed mesh-based method. Even though MATLAB is used for both of the methods, one needs to consider that the pixel-based approach is using the internal optimized DCT function (written in C) to solve the linear system at each iteration, unlike the mesh-based technique which uses not-optimized MATLAB functions written by the authors. However the mesh-based technique outperforms the pixel-based method both in accuracy and speed. Higher speeds can be achieved with C implementations of the proposed method. Usually image registration techniques, specially the non-parametric ones, work on the pixel level. Multi-resolution techniques do not distinguish between regions that have significant feature content and regions that are featureless/uniform. Octrees are a way of to adaptively sub-divide images based on feature content. However, the rectangular  boundaries in octrees do no suit feature boundaries that tend to be curvilinear. On the other hand triangle meshes can accurately follow curvilinear feature boundaries. However using triangular mesh has its own implications regarding defining  the problem of image registration and optimization of the geometric transformation needed to be applied to template image to match the reference image. Here a new technique for fast mesh-based image registration is proposed which can take into account these implications and achieve high accuracy. This method has the dual advantage of a compact representation and fast computation. Furthermore, images at any desired resolution can be considered for registration since we only need to deal with the mesh nodes and not image pixels.

\end{document}